\documentclass[conference]{IEEEtran}
\IEEEoverridecommandlockouts
\usepackage{cite}
\usepackage{amsmath,amssymb,amsfonts}
\usepackage{algorithmic}
\usepackage{graphicx}
\usepackage{stfloats}
\usepackage{textcomp}
\usepackage{xcolor}
\def\BibTeX{{\rm B\kern-.05em{\sc i\kern-.025em b}\kern-.08em
    T\kern-.1667em\lower.7ex\hbox{E}\kern-.125emX}}

\begin{document}
\title{{Deepfake Detection with Spatio-Temporal Consistency and Attention}
}

\author{

\IEEEauthorblockN{Yunzhuo Chen}
\IEEEauthorblockA{\textit{The University of Western Australia} \\
Perth, Australia \\
yunzhuo.chen@research.uwa.edu.au}
\and

\IEEEauthorblockN{Naveed Akhtar}
\IEEEauthorblockA{\textit{The University of Western Australia} \\
Perth, Australia \\
naveed.akhtar@uwa.edu.au}
\\
\IEEEauthorblockN{Ajmal Mian}
\IEEEauthorblockA{\textit{The University of Western Australia} \\
Perth, Australia \\
ajmal.mian@uwa.edu.au} 
\and

\IEEEauthorblockN{Nur Al Hasan Haldar}
\IEEEauthorblockA{\textit{The University of Western Australia} \\
Perth, Australia \\
nur.haldar@uwa.edu.au}

}

\maketitle

\begin{abstract}
Deepfake videos are causing growing concerns among communities due to their ever-increasing realism. Naturally, automated detection of forged Deepfake videos is attracting a proportional amount of  interest of researchers. Current methods for detecting forged videos mainly rely on global frame features and under-utilize the spatio-temporal inconsistencies found in the manipulated videos. Moreover, they fail to attend to manipulation-specific subtle and well-localized pattern variations along both spatial and temporal dimensions. Addressing these gaps, we propose a neural Deepfake detector that focuses on the localized manipulative signatures of the forged videos at individual frame level as well as frame sequence level. Using a ResNet backbone, it strengthens the shallow frame-level feature learning with a spatial attention mechanism. The spatial stream of the model is further helped by fusing texture enhanced shallow features with the deeper features. Simultaneously, the model processes frame sequences with a distance attention mechanism that further allows fusion of temporal attention maps with the learned features at the deeper layers. The overall model is trained to detect forged content as a classifier. We evaluate our method on two popular large data sets and achieve significant performance over the state-of-the-art methods. 
Moreover, our technique also provides memory and computational advantages over the competitive techniques.   

\end{abstract}


\section{Introduction}

Advancements in generative models has made forged visual content more realistic, creating difficulties in distinguishing between real and fake visual content. 
Various facial image forgery techniques are being continuously proposed and improved\cite{kemelmacher2016transfiguring,thies2016face2face,koujan2020head2head,nirkin2019fsgan,pumarola2018ganimation,wu2018reenactgan,natsume2018rsgan,suwajanakorn2017synthesizing}. The forged content created by the Deepfake\cite{westerlund2019emergence} face-changing technology first appeared on the community forum Reddit\cite{roettgers2018porn}. Since then, it has set off a wave of creating fake content\cite{bitesizedeepfakes}. 
Deepfake generative models are fast approaching the precision that can fool even the human visual system.
Unfortunately, software and source code for Deepfake content generation are rather easy to access. This convenience  of making fake videos and pictures has led to a wide  abuse of this technology, resulting in serious law infringements, e.g., using this technology to create obscene or pornographic content\cite{roettgers2018porn}. Deepfake technology can also create images and videos of political  figures to spread rumors that can have serious consequences. 

\begin{figure}[t]
\centering
\includegraphics[ width= 0.45\textwidth]{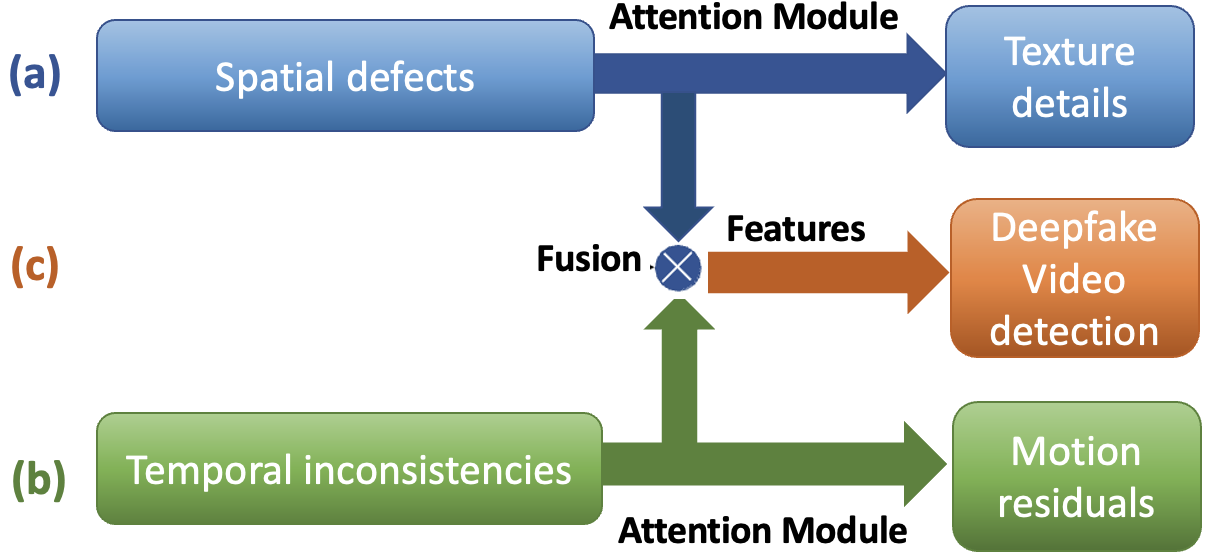}
\caption{The proposed method leverages three major components. (a) Attention mechanism in the spatial domain to capture Deepfake related spatial artifacts appearing in individual frames. (b) A temporal attention module that captures temporal inconsistencies between the consecutive frames. (c) A fusion mechanism followed by the detection stage to make the prediction.}
\end{figure}

To mitigate the threat posed by Deepfake technology, several Deepfake detection techniques have been  proposed~\cite{rossler2019faceforensics++,li2020face,li2018ictu,masi2020two,wu2020sstnet}. Majority of these techniques solve a binary classification problem using global features of images obtained through a backbone network, and processed by the classifier. 
However, due to the advancement of Deepfake generative techniques, such a classical detection paradigm now faces diverse challenges. For instance, with the Deepfake advancement, the difference between the forged content and genuine visual content is becoming increasingly subtle and well-localized.
This makes discriminative encoding of the suspect Deepfake features in a holistic manner very challenging. Moreover, the classical paradigm also suffers from leveraging content features sub-optimally.  Insufficient utilization of the visual features adversely affects the detection performance. Hence, most detection methods based on the global features perform unsatisfactory. Detecting the subtle changes in fake videos is more closely related to the fine-grained classification task~\cite{zhao2021multi}. Therefore, this work approaches the problem from the  fine-grained classification perspective, thereby, enabling effective detection of the subtle and local Deepfake signatures in the forged visual content. 

\begin{figure*}[t]
\centering
\includegraphics[ width=\textwidth]{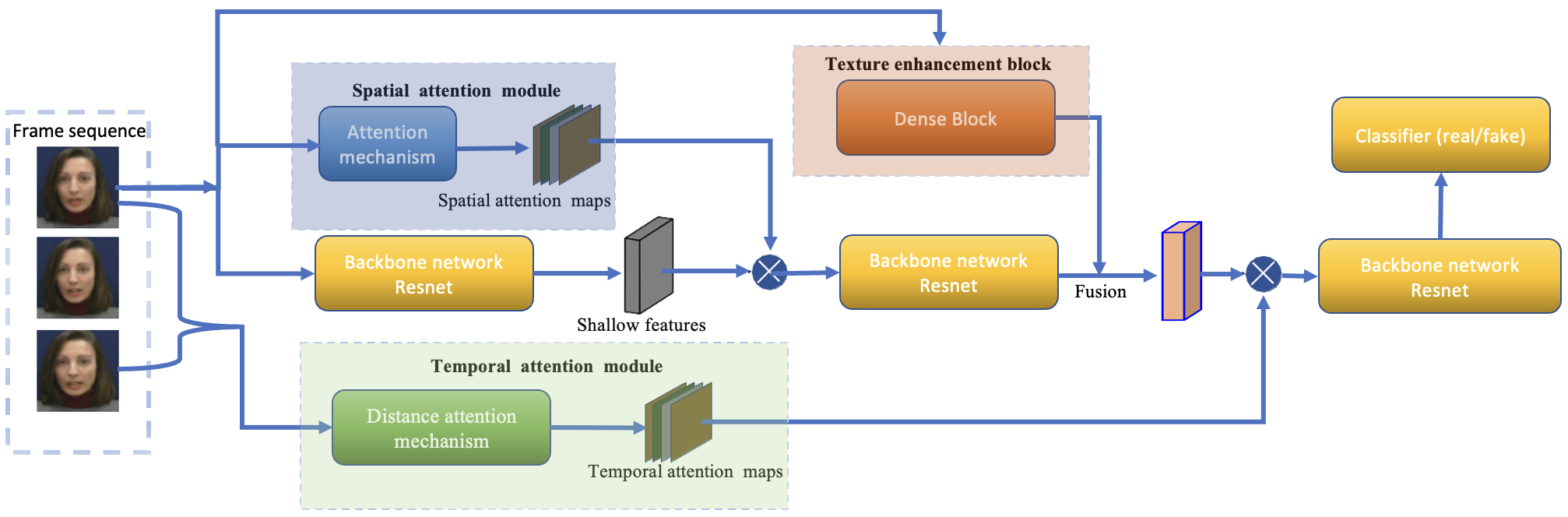}
\caption{The framework of our method consists of three important components. A texture enhancement block module for enchancing the texture features. A spatial attention module to capture Deepfake related spatial artifacts appearing in individual frames. A temporal attention module captures inconsistencies between consecutive frames. Guided by the three components, the backbone network can focus on local regions in the detection task.} 
\label{fig:framework}
\end{figure*}

We propose a Deepfake video detection technique that focuses on the consistency of spatio-temporal features of subtle expressions. During feature construction,  changes between the frames can reflect more fine-grained correlation information. Therefore, we consider two consecutive frames in the video as the input data objects and calculate the optical flow between the frame pair. In sequential image frames, optical flow~\cite{horn1981determining} can be defined as the distribution of apparent velocities of movement of the brightness pattern. Our model uses the convolutional neural network (CNN)~\cite{o2015introduction} to extract spatial domain features. Further, we use the optical flow method to describe Deepfake time-domain features. Finally, the model fuses the temporal and spatial features to perform classification for the Deepfake detection.

Neural networks are known to learn effective representations that can be used to identify important data features. Recent advances in the attention mechanism for neural networks~\cite{vaswani2017attention} allow focusing on the most relevant features for a given task.
Our method exploits the attention mechanism for both spatial and temporal content in videos to let our model concentrate on the subtle changes in the data that are peculiar to fake content. 
With an additional use of texture enhancement and employing a strong backbone for  feature encoding, our technique is able to catch  subtle changes between real and fake content.   
We have carried-out extensive experiments on two publicly available datasets,  FaceForensics++(LQ)~\cite{rossler2019faceforensics++} and DFDC ~\cite{dolhansky2020deepfake}, where the proposed model performs better than eight state-of-the-art methods. The contributions of this paper are as follows: (1) We undertake Deepfake detection of videos as a fine-grained classification task. (2) We propose a detection method that not only uses attention mechanism for the spatial content, but also along the temporal dimension to identify Deepfake related inconsistencies in subtle facial expressions. (3) We verify the effectiveness of our technique through extensive comparative experiments and ablation study.

\section{RELATED WORK}

Some of the early Deepfake detection techniques \cite{yang2019exposing, liy2018exposingaicreated} also used visual biological artifacts to identify the authenticity of visual content. More recent methods  rely on  learning-based methods. Li et al.\cite{li2021exposing} used several classic high-complexity deep neural networks to extract more detailed features from the forged videos and used them for the underlying classification task. Wu et al.~\cite{wu2020sstnet} proposed that spatial and temporal features can be used as detection targets. They model the temporal dimension with LSTM\cite{6795963} and employed  Xception\cite{chollet2017xception} for the convolutional network. Wu et al.~\cite{wu2020sstnet} proposed that spatial and temporal features can be used as detection targets. They model the temporal dimension with LSTM\cite{6795963} and employed  Xception\cite{chollet2017xception} for the convolutional network.However, highly complex deep neural networks and excessive iterative training can lead to overfitting of their method to the training data.

\section{PROPOSED METHOD}
\begin{figure}[t]
\centering
\includegraphics[ width=8cm]{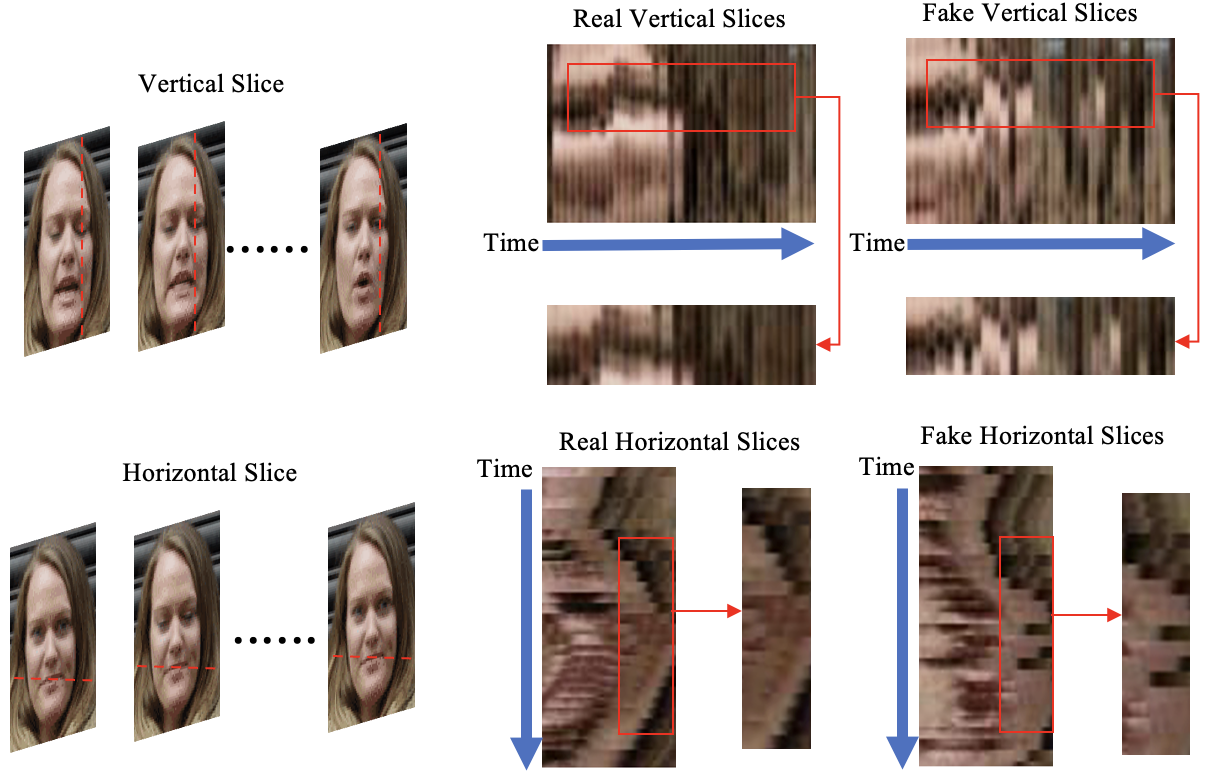}
\caption{Temporal disparity between real and fake videos. Motion at a certain  position of the video is visualized in vertical and horizontal slices. The fake video slices are far less smooth than the real ones. The red dotted line is the position of the slice.} 
\label{fig:temporal}
\end{figure}

The schematic of the proposed technique is given in Fig.~\ref{fig:framework}. However, before discussing the details of the method, we first give the motivation behind the use of spatio-temporal information in our technique.  We select a pair of real and fake videos with small differences, and then divide them into 100 frames, respectively. In Fig.~\ref{fig:temporal},  vertical slices and horizontal slices of a real video and its fake counterpart are plotted. It is apparent that there are discernible differences between the two. The obvious differences are in the flickering and  jittering patterns. The reason is that,  Deepfake frames generally do not account for the context from  the previous frames, and process each frame individually to meet the underlying manipulation objective. Since Deepfake videos are often generated frame by frame, there exist some inconsistencies between the frame sequences. To fully capitalize on such inconsistencies, our method employs different modules for extracting texture features and capturing spatial and temporal defects. The model contains two attention mechanisms that can be used to combine the information to better focus on the  suspected local regions.

Zhao et al.~\cite{zhao2021multi} found that the artifacts caused by generative techniques are mainly retained in the texture information of the shallow features. To effectively detect artifacts in shallow features, we incorporate a texture enhancement block in the spatial module to extract local texture features. The spatiotemporal inconsistency between frames is mainly reflected in high-level semantic features. We use an attention mechanism to guide high-level semantic features in the temporal attention module. Based on the above insights, our framework in Fig.~\ref{fig:framework} consists of three major modules supporting the backbone, i.e., texture enhancement block, spatial attention module and temporal attention module. 

We extract and enhance texture features from the input data using a Dense block~\cite{Huang_2017_CVPR}. The spatial attention module captures the spatial artifacts while paying attention to the shallow features. The temporal attention module is used to capture temporal inconsistencies between frames. Our model processes feature maps through attention modules in both temporal and spatial domains, and makes the model focus on key local regions. We use ResNet50~\cite{He_2016_CVPR}, a well known CNN, as the backbone in our  model. Below, we discuss each module individually.

\subsection{ Texture enhancement block}

This module is used to enhance the textual features available in the video frames. We use a standard Dense block\cite{song2020efficient}, which employs the  dense connection mechanism. This means that every layer is connected to all subsequent layers.  Specifically, each layer  accepts all previous layer outputs as its input. This mechanism is known to learn a more effective representation as compared to the a simple feedforward connection  mechanism. 
The main function of the Dense block in our model is to extract and enhance texture information from shallow feature maps.

\subsection{Spatial attention module}\label{AA} 
The main purpose of the spatial attention module is to capture the artifacts present in the spatial domain. Since the Deepfake generation network must operate on each frame, it can introduce unnatural facial structures and mismatched texture details in the individual frames. Correlating within frame information helps to detect such inconsistencies. Using a pooling mechanism to collect frame information is the most common method~\cite{hu2018squeeze}. However, encoding discernible global information is hardly effective when forged artifacts are localized. To capture artifacts in the spatial domain, our backbone network has a module that adds attention. The module is the Weakly Supervised Data Augmentation Network (WS-DAN)~\cite{hu2019see} based on Resnet50 as the backbone. WS-DAN addresses the classification problem from a fine-grained perspective by building an attention map to extract local features.

The attention mechanism based on fine-grained detection can solve the problems of oversampling and insufficient texture information in local regions. Since we want to also capture defects in the spatial domain from a global perspective using fine-grained attention mechanism, operating on individual frames of the video is imperative. The feature map of the backbone network is constructed from the attention map generated by a single frame. Such a mapping strategy is useful for re-calibrating the importance of  regions with respect to the overall objective of the detection task. Since the texture features are dominant in the shallow features, we apply attention early in our model. WS-DAN performs feature extraction on the frame and then obtains attention maps through convolution operations. The attention map generated by this mechanism is multiplied element-wise with a shallow feature map, which results in a feature map that emphasizes the local regions for further processing by the backbone.

\subsection{Temporal attention model}

\begin{figure}[t]
\centering
\includegraphics[ width=9cm]{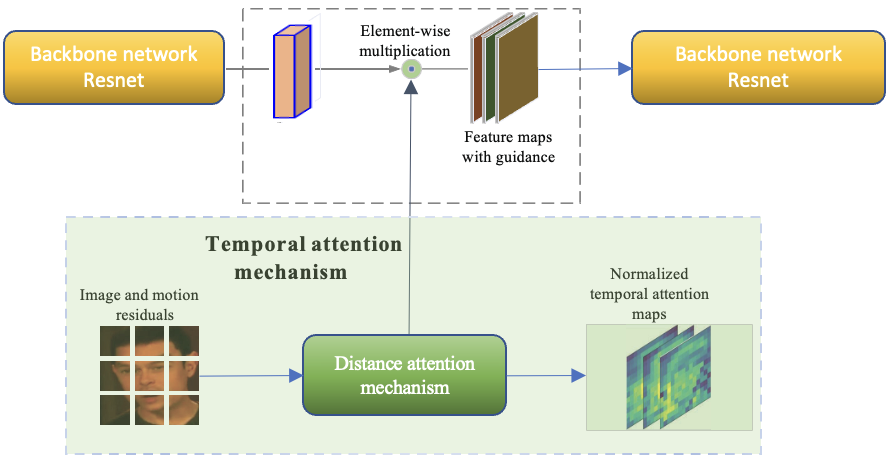}
\caption{Temporal attention incorporation in the backbone. A single frame and its motion residuals are taken as input. The generated attention map is used to guide the feature map.} 
\label{fig:temp1}
\end{figure}

Temporal features mainly characterize the differences in facial expressions of characters over time, such as opening mouth, smiling, blinking and other subtle movements. Since Deepfake videos are generally produced frame-by-frame, the subtle changes in these facial expressions are incoherent and inconsistent. Therefore, it is very relevant for DeepFake detection  to analyze the related temporal features from the video frames.

We use consecutive frames to capture temporal inconsistencies. The attention module in the temporal domain needs to use the next $n$ frames to detect the target frame. Details on selecting the value of $n$ are given in  Section~\ref{sec:experiment}. For temporal attention, we are more concerned with the changes of the video in the temporal dimension. Therefore,  motion residuals between adjacent frames are used as the input, as depicted in Fig.~\ref{fig:temp1}. Optical flow~\cite{horn1981determining} of the motion residual describes the change trend and motion amplitude of the frames.  From the perspective of physical meanings, optical flow  describes the relationship between the objects in the video frames along the time dimension, thereby establishing the relationship between the  consecutive images in the video.
This makes it a desired tool to analyze the temporal inconsistencies in the forged videos. Hence, this is what we use as the underlying method for the temporal stream of our model. 

To apply attention, we need to compute the motion residuals. The process of calculating those is as follows.
First, we define the function $I(x, y, t)$ at time $t$ at position $(x, y)$ in the image. After time $dt$, the pixel moves by $(dx,dy)$ in the next frame. Since the pixels are the same, we can write
\begin{equation}
    I(x,y,t)=I(x+dx,y+dy,t+dt).
    \label{equ:func}
\end{equation}
\noindent Applying the first-order Taylor expansion to R.H.S.~of Eq.~(\ref{equ:func}) 
\begin{equation}
\begin{split}
    I(x+dx,y+dy,t+dt) = I(x,y,z)+\\(\delta I/\delta x)dx+(\delta I/\delta y)dy+(\delta I/\delta t)dt+\xi.
\end{split}
\label{equ:taylor}
\end{equation}
\noindent Here, $\xi$ refers to the second-order infinitesimals in the Taylor expansion. Ignoring $\xi$ and bring Eq.~(\ref{equ:func}) into Eq.~(\ref{equ:taylor}), we have 

\begin{equation}
    (\delta I/\delta x)dx+(\delta I/\delta y)dy+(\delta I/\delta t)dt = 0.
\end{equation}
Divide both sides of Equation (3) by $dt$ we have
\begin{equation}
    (\delta I/\delta x)(dx/dt)+(\delta I/\delta y)(dy/dt)+(\delta I/\delta t) = 0.
\end{equation}

The above module is used for motion residual computation. For consecutive frames, the optical flow  is a vector field encoding apparent motion. The motion vector behaves  differently  for the tampered frames and the real frames. The motion field of the real frame is expected to be  generally smooth and coherent. However, the motion vector of the tampered locations (such as eyes or mouth) may appear relatively incoherent. This difference is a manifestation of the spatio-temporal inconsistency between real and fake frames. Our model uses optical flow technique to capture Deepfake time-domain inconsistensies.

\begin{figure*}[t]
\centering
\includegraphics[width=18cm]{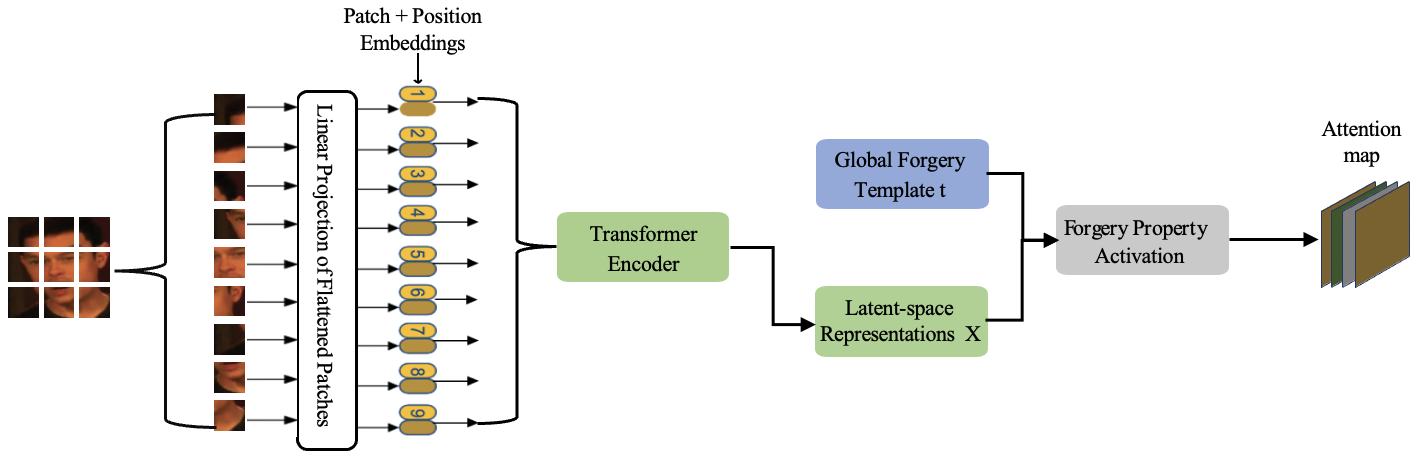}
\caption{The employed long-distance attention mechanism. This mechanism divides the input into patches and treats these patches as a sequence(3×3). The patches in the sequence are converted into vectors and then formed into matrices with their dimensions. The matrix transforms the patch embeddings into the latent space. Finally, a global forgery template obtains attention weights from the latent space (X) to generate the attention map.} 
\label{fig:longDist}
\end{figure*}

Our temporal attention utilizes a ViT based distance attention mechanism \cite{6786458, lu2021detection}, that is illustrated in  Fig.~\ref{fig:longDist}. This mechanism divides the input into patches and treats these patches as a sequence. The sequence length is 9 (3×3) in our model. Then, it flattens each patch in the sequence, and converts each patch to a vector. Assume the dimension of each vector is $C$. Then, the input becomes a matrix in $\mathbb R^{9 \times C}$. This matrix is fed to the Transformer Encoder along with a vector of class tokens\cite{jiang2021all} to process image patch sequences. In the temporal attention module, the inconsistency between frames can be modeled by a global fake template. The activations obtained by the global fake template from the latent space (X) are the attention weights.  The attention map generated by the temporal attention module is used to guide feature maps in the temporal domain, which are then fused with features in the spatial domain. The fused features are fed into the classifier to get the final result.

\section{EXPERIMENTS}
\label{sec:experiment}
We present extensive experimental results to demonstrate the effectiveness of our method. 

\subsection{Datasets} 
Two mainstream Deepfake datasets FaceForensics++ (FF++)~\cite{rossler2019faceforensics++} and Deepfake Detection Challenge (DFDC)~\cite{dolhansky2020deepfake} are used in our experiments. Advances in Deepfake generative models have resulted in  high video quality in both datasets. 
FF++ is a large-scale dataset and multiple generative techniques were used in the production process. Most of the videos have faces that are not occluded and can be tracked. Each video in the dataset has three versions, namely RAW, high quality (HQ) and low quality (LQ). DFDC has a data volume of up to 472GB, including 119,197 videos. Each video is 10 seconds long, but the frame rate varies from 15 to 30fps, and the resolution varies from 320x240 to 3840x2160. Among the training videos, 19,197 videos are real footage of about 430 actors, and the remaining 100,000 videos are fake videos generated from real videos.

\subsection{Evaluation metric}

As a binary classification problem, we use Accuracy (ACC)\cite{matern2019exploiting} metric to evaluate the performance of our proposed model. A higher value in ACC justifies better performance of a model. 
We also use Area Under the ROC Curve (AUC)  as another evaluation metric. The ROC curve sorts the samples according to the size of the prediction results of the model. The predicted probability of each sample is used as the threshold value. 
The ROC curve is calculated with false positive rate as the horizontal axis and true positive rate as the vertical axis curve. The ROC curve can better describe the generalization performance of the model. The larger the AUC, the better the performance of the model. 

\subsection{Experimental setup and parameters}
We use Dlib\cite{king2009dlib} to process the dataset. The Dlib face detector can detect a face and return the corresponding rectangular box. The face sizes in the backbone network and attention mechanism are 398×398 and 224×224, respectively. 
The iteration number of the model is set to $80$ and the loss function of the deep learning model is set to mean-square error (MSE). 

The experimental dataset is divided into a training set and a validation set with a ratio $80:20$.

As the number of training iterations increases, model learning becomes better. However, due to the strong fitting ability of  deep learning models, it is  easy to overfit the training data. To avoid overfitting, an early stopping strategy is adopted in our experiments. Such a strategy is often used in the training of deep learning models. Therefore, when the loss function value of the model does not improve over a period of time, the training of the model is terminated. 

\subsection{Result and analysis}
\label{subsec:result}

First, in Table~\ref{tab:confusionMatrix1}, we report the confusion matrix of our evaluation on  FF++(LQ).  
As can be seen from the table, our model performs reasonably well in terms of correctly predicting the positive and negative samples.

According to the shown confusion matrix, the detection accuracy of the model is 98.1$\%$, the recall rate is 97.14$\%$, 
and the true positive and true negative prediction accuracies are 98.99$\%$ and 97.24$\%$, respectively.

\begin{tiny}
\begin{table}[t]
\scriptsize
\centering
\renewcommand\arraystretch{1.1}
\setlength{\tabcolsep}{7mm}{
\caption{Confusion matrix of the prediction results on the FF++(LQ) validation set.}
\label{tab:confusionMatrix1}
\begin{tabular}{lcccccccccc}
\hline
Real tag & Prediction result & Prediction result \\
 &  (positive) &  (negative)\\
\hline
Positive & 1869 & 55 \\
Negative & 19 & 1943 \\
\hline
\end{tabular}}
\end{table}
\end{tiny}

We further compare our method with eight state-of-the-art techniques using ACC and AUC metrics in Table~\ref{tab:resultCompair} on FF++ dataset. Since some existing reported results are difficult to reproduce, we compare only with reproducible models. To clearly compare the memory and computational requirements for different methods, we also report the memory and FLOPS for the methods in Table~\ref{tab:resultCompair}, whenever possible. 
The AUC values not only reflect the detection effectiveness, but also indicate the generalization ability of the model. 
In the overall results of the experiment in Table~\ref{tab:resultCompair}, the detection performance  of the proposed model is consistently better in both AUC and ACC metrics. Moreover, the memory requirements and computational efficacy also appear to be more desirable for our method. 

\begin{tiny}
\begin{table}[t]
\scriptsize
\centering
\renewcommand\arraystretch{1.1}
\setlength{\tabcolsep}{1.2mm}{
\caption{Results on the FF++(LQ) dataset. The results for \cite{afchar2018mesonet}, \cite{chollet2017xception} and \cite{bayar2016deep} are from~\cite{rossler2019faceforensics++}. The results for\cite{rossler2019faceforensics++} and \cite{nguyen2019multi} are from \cite{tolosana2020deepfakes}}
\label{tab:resultCompair}
\begin{tabular}{lcccrr}
\hline
Model & ACC$\%$ & AUC &  Param.(MB) & Memory(MB) & MFLOPS\\
\hline
Cozzolino et al. \cite{rossler2019faceforensics++} & 85.23 & 0.9120 & - & - & - \\
Bayar et al. \cite{bayar2016deep} & 66.84 & - & - & - & - \\
Xception \cite{chollet2017xception} & 81.00 & 0.6653 & - & - & - \\
Capsule-forensics \cite{nguyen2019capsule} & 83.33 & - & - & - & - \\
C3D \cite{tran2015learning} & 78.67 & 0.8400 & 61.10 & \textbf{4.19} & \textbf{716}\\
Resnet \cite{he2016deep} & 86.53 & 0.9780 & 60.19 & 226.06 & 11570\\ 
ClassNseg \cite{nguyen2019multi} & 74.64 & 0.8173 & 88.79  & 276.02 & 16490\\ 
MesoNet \cite{afchar2018mesonet} & 70.47 & 0.5360 & 44.55 & 161.75 & 7840\\
\hline
Our model & \textbf{90.91} & \textbf{0.9810} & \textbf{25.56} & 109.69 & 4120\\ 

\hline
\end{tabular}}
\end{table}
\end{tiny}

\subsection{Cross-dataset performance}
We further evaluate the generalization  of our framework in this section. 
To that end, cross-dataset results of our technique and comparison to existing methods are provided in Table~\ref{tab:crossDataset}. 
The second column in Table~\ref{tab:crossDataset} shows the results of models trained on FaceForensics++ (LQ) and tested on FaceForensics++ (LQ). The third column in Table~\ref{tab:crossDataset} shows the results of the models trained on FaceForensics++ (LQ) and tested on DFDC.
Most Deepfake detection models often lose some
performance when tested across datasets. The main reason for this is the variance between datasets. For example, the video resolution in DFDC is much higher than in FF++(LQ). According to the statistics of ~\cite{rossler2019faceforensics++}, the average accuracy of most mainstream deepfake detection model in FF++ (LQ) is 58.7$\%$. This proves that our model can effectively extract features when faced with highly compressed datasets.

In the shown results, Xception only uses the Xception model trained for the DeepFake detection task. It can be seen as a backbone network for this task. From this viewpoint,  our results confirm that our spatial-temporal modelling can effectively emphasize on the local regions, thereby considerably improving the performance.

\begin{table}[t]
\begin{tiny}
\scriptsize
\centering
\renewcommand\arraystretch{1.1}
\setlength{\tabcolsep}{5mm}{
\caption{Results  (ACC$\%$) when models are trained on the FF++ dataset and tested on FF++ test set and the DFDC dataset.}
\label{tab:crossDataset}
\begin{center}
\begin{tabular}{lcccccccccc}
\hline
Model  & FF++(LQ) & DFDC \\
 & Test Results & Test Results \\
\hline
MesoNet\cite{afchar2018mesonet} & 70.47 & 84.70 \\
Xception\cite{chollet2017xception}& 90.03&89.70\\
FWA\cite{li2018exposing}& 80.10&56.90\\
DSP-FWA\cite{li2019exposing}  & 93.00& 64.60\\
Two Branch\cite{masi2020two} & 93.18& 73.41 \\
\hline
Our model & \textbf{90.91}& \textbf{95.97}\\
\hline
\end{tabular}
\end{center}}

\end{tiny}
\end{table}

\subsection{Discussion}

\begin{figure}[t]
\centering
\includegraphics[ width=8.5cm]{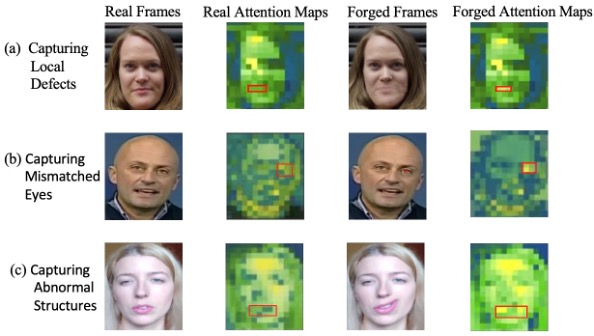}
\caption{Attention maps generated by single frames.}
\label{fig:spatAtt}
\end{figure}

It is difficult for the human eye to observe the difference between attention maps generated from real and fake visual content. Therefore, to further observe the effect of the attention mechanism we manually add exaggerated defects to the real frames. In Fig.~\ref{fig:spatAtt}, the first and third columns consist of real frames and tampered fake frames, respectively. The second and fourth columns are the attention maps generated by the real and fake frames, respectively. We use red boxes to mark the difference between real and fake attention maps. Fig.~\ref{fig:spatAtt} indicate that the tampered positions in the face are correctly marked.

\begin{figure}[t]
\centering
\includegraphics[ width=8.5cm]{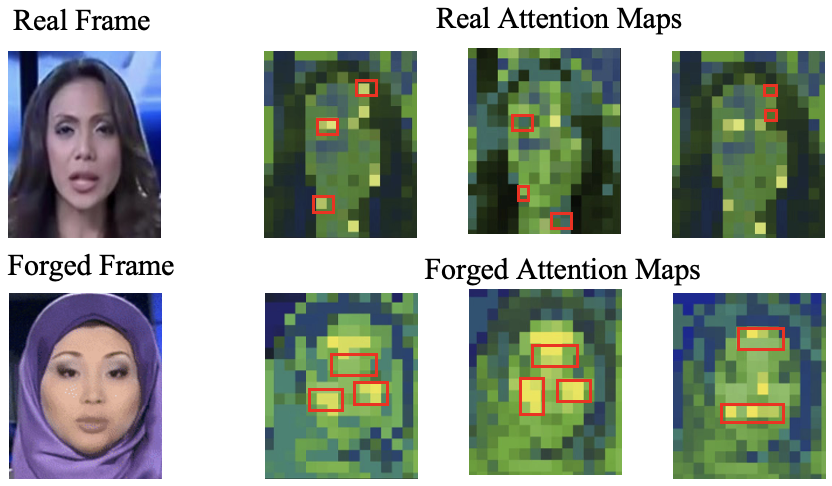}
\caption{Attention maps generated by consecutive frames.}
\label{fig:temAtt}
\end{figure}

In Fig.~\ref{fig:temAtt}, the attention changes between three consecutive frames are marked with red boxes. The attention map generated from the real video is smoother, which means differences between real attention maps are few and evenly dispersed. On the other hand, tampered frames often have issues with jitter and flicker, which leads to the uneven brightness of the temporal attention maps produced by fake videos. Comparing with the real attention maps, the fake attention maps are larger and have more concentrated bright regions. Therefore, Fig.~\ref{fig:spatAtt} and Fig.~\ref{fig:temAtt} together indicate that the used attention mechanism can capture the defects in local and global perspectives, and make the backbone network focus more on the pivotal regions.

\section{Conclusion}

Current Deepfake detection methods are generally based on global and spatial content features and ignore the importance of time-domain and localized features. 
Therefore, to leverage  the subtle incoherence and inconsistency of facial expression changes in face-changing forged content, we devise a novel spatio-temporal technique that uses attention mechanism to account for subtle variations. 
Our method uses a convolutional neural network  to extract spatial domain features of Deepfake images and uses an  optical flow technique to describe Deepfake time-domain features.
The backbone of our model uses these features after finding the localized regions that require more attention to enable better performance. Finally, we fuse the temporal and spatial features of the content and treat the detection as a binary classification task. Extensive results on two large datasets establish that the proposed method not only has a clear performance gain over the related existing methods, but also has memory and computational advantages. 

\section{ACKNOWLEDGMENTS}
This research was in part supported by the UWA Univesity Postgraduate Award (UPA) and by ARC Discovery Grant
DP190102443. Dr. Naveed Akhtar is a recipient of
Office of National Intelligence National Intelligence Postdoctoral Grant (project number NIPG-2021-001) funded by the Australian Government. Professor Ajmal Mian is the recipient of an Australian Research Council Future Fellowship Award (project number FT210100268) funded by the Australian Government.

{\small
\bibliographystyle{ieee_fullname}
\bibliography{egbib}

\end{document}